# Fine-Tuning Small Language Models (SLMs) for Autonomous Web-based Geographical Information Systems (AWebGIS)

Mahdi Nazari Ashani[1], Ali Asghar Alesheikh[1, 2, *], Saba Kazemi[1], Kimya Kheirkhah[1], Yasin Mohammadi[1], Fatemeh Rezaie[1], Amir Mahdi Manafi[3], Hedieh Zarkesh[1]

[1] Department of Geospatial Information Systems, Faculty of Geodesy and Geomatics Engineering, K. N. Toosi University of Technology, Tehran, Iran

[2] Center of Excellence for Geospatial Information Technology, Faculty of Geomatics Engineering, K. N. Toosi University of Technology, Tehran, Iran

[3] Department of Computer and Information Technology, Shahr-e-Rey Branch, Islamic Azad University, Tehran, Iran

* alesheikh@kntu.ac.ir

## Abstract

Autonomous web-based geographical information systems (AWebGIS) aim to perform geospatial operations from natural language input, providing intuitive, intelligent, and hands-free interaction. However, most current solutions rely on cloud-based large language models (LLMs), which require continuous internet access and raise users' privacy and scalability issues due to centralized server processing. This study compares three approaches to enabling AWebGIS: (1) a fully-automated online method using cloud-based LLMs (e.g., Cohere); (2) a semi-automated offline method using classical machine learning classifiers such as support vector machine and random forest; and (3) a fully autonomous offline (client-side) method based on a fine-tuned small language model (SLM), specifically T5-small model, executed in the client's web browser. The third approach, which leverages SLMs, achieved the highest accuracy among all methods, with an exact matching accuracy of 0.93, Levenshtein similarity of 0.99, and recall-oriented understudy for gisting evaluation ROUGE-1 and ROUGE-L scores of 0.98. Crucially, this client-side computation strategy reduces the load on backend servers by offloading processing to the user's device, eliminating the need for server-based inference. These results highlight the feasibility of browser-executable models for AWebGIS solutions.

## 1. Introduction

Recent studies highlight that large language models (LLMs), such as ChatGPT, have significantly advanced natural language processing (NLP) through their development and application across diverse language-based tasks. Human language comprehension and generation has been revolutionized by these models, enabling machines to achieve advanced proficiency in understanding, reasoning about, and producing text across diverse applications (Zubiaga 2024). Concurrently, integrating NLP techniques with domain-specific systems has created new opportunities to automate complex processes (Guan et al. 2024). One of the notable domains in which NLP is modifying its traditional workflows is geospatial information systems (Akinboyewa et al. 2025).

Autonomous geographical information systems (GIS), also known as language-driven GIS, refers to the application of LLMs in this field, resulting in systems able to process spatial queries, generate and visualize maps, and use natural language to automate data transformation (Akinboyewa et al. 2025; Li and Ning 2023; Y. Zhang, Li, et al. 2024; Y. Zhang, He, et al. 2024). A study conducted by (Li and Ning 2023) showcases smart systems based on autonomous GIS which are independent of constant user input and perform a wide variety of geospatial tasks. By expediting the pace of repetitive operations such as querying spatial databases, applying spatial filters, and generating thematic maps, these systems enhance performance and productivity in academic research and industrial applications.

According to (Li and Ning 2023), autonomous GIS is capable of more than just carrying out basic commands. These systems can comprehend and carry out complex workflows at an increasing rate by utilizing only natural language, such as chaining multiple spatial operations, conducting analyses, or dynamically modifying parameters (Y. Zhang, Li, et al. 2024). As a result of this paradigm shift, user interaction with GIS is evolving, becoming more effective, inclusive, and user-friendly.

Despite these advancements, significant challenges remain in the practical deployment of language-driven systems. Most LLMs require high computational resources, making them unsuitable for low-power devices like smartphones, tablets, or embedded IoT platforms (Shen et al. 2025; Q. Zhang, Liu, and Pan 2025). Relying on centralized processing—where queries and data are sent to cloud servers—introduces vulnerabilities and increases the risk of data breaches (Lorestani, Ranbaduge, and Rakotoarivelo 2024). While privacy-preserving technologies such as federated learning or blockchain-based data storage have been proposed as solutions (Belal et al. 2024; Farnaghi and Mansourian 2020; Rao et al. 2021), implementing fully decentralized autonomous GIS systems remains a technical challenge—especially in smaller, resource-constrained devices. Due to these challenges, it is crucial to look for alternative architectures that balance automation capabilities with practical limitations, such as latency and computational restrictions. To tackle this necessity, our study conducts a comparative evaluation of three different approaches for implementing language-driven GIS automation:

1. Fully automated online approach: few-shot learning with cloud-based LLMs, namely GPT or Gemini, is utilized in this method to analyze and execute spatial queries dynamically.

Automation and flexibility are maximized in this approach; however, it heavily relies on internet connectivity and raises serious concerns about computational costs.

2. Semi-automated offline approach: To classify prompts, this approach makes use of machine learning classifiers, such as decision trees and support vector machine (SVM). Even though there is an absence of some capabilities such as argument detection and full command generation, it allows basic automated functionality independent from internet connection. a fine-tuned small language model

3. Fully automated offline approach (our proposed method): In this approach, a fine-tuned small language model (SLM), including T5-small model, is implemented to understand user inputs and produce GIS function calls with appropriate parameters. It successfully tackles issues related to latency and hardware limitations due to its entirely-offline nature. For this reason, it is suitable for deployment on edge devices such as mobile phones and tablets, even with limited internet access within real-life context.

Each of these approaches is evaluated against key criteria including automation level, internet dependency, computational efficiency, and accuracy. Through this comparative study, our goal is to identify the most suitable strategies for deploying natural-language-powered GIS systems across a variety of technical and operational contexts—from high-performance cloud systems to lightweight tools.

## 2. Related work

Artificial intelligent has experienced a really big jump with the advent of LLMs in recent years (Chang et al. 2024). Transformers that were introduced in 2017 (Vaswani et al. 2017) were a starting point to make LLMs based on them by release of BERT (Devlin et al. 2019), GPT-2 (Radford et al. 2019), T5-small model (Raffel et al. 2020), GPT-3 (Brown et al. 2020), and so on up to 2020. It was November, 2022 that the first ChatGPT was released and made a huge impact in AI industry due to its capability versus previous models (Cong-Lem, Soyoof, and Tsering 2025). Later, other models with better performance were released by different companies and institutions such as LLaMA (Touvron et al. 2023), GPT-4 (Achiam et al. 2023), Gemini (G. Team et al. 2024), DeepSeek (C. Zhang et al. 2025), and so on. These models have made a huge effect on the growth of AI based systems.

For instance, the introduction of models such as GPT-3 represented a significant advancement in the field of LLMs, revealing how massive pretraining on diverse datasets helps with enhancing few-shot learning and emerging cross-domain capabilities (Brown et al. 2020). On top of that, the GPT-4 technical document demonstrates progress in logical reasoning, multilingual interpretation, and alignment with safety standards (Achiam et al. 2023). Likewise, the LLaMA series has offered more compact yet robust alternatives that advance the frontier of open-access LLM development, facilitating wider academic and industrial involvement (Touvron et al. 2023).

The combination of LLMs with GIS is enabling new forms of spatial intelligence and user interaction (S. Wang et al. 2024). Natural language interfaces in GIS enable users to interact with complex spatial datasets using everyday language, making the technology more accessible to non-experts (Li, Grossman, et al. 2025). This innovation is built upon early research in parsing natural language to structured geospatial data querying (Ning et al. 2025). Additionally, recent developments have focused on fully automated GIS operations, advancing towards autonomous GIS systems (Ning et al. 2025). MapColorAI leverages LLMs to design context-aware choropleth map color schemes that align with user intent and data semantics (Yang et al. 2025). This system demonstrates how LLMs can enhance personalization and usability in GIS tasks that traditionally required expert knowledge. These frameworks use LLMs to interpret instructions and spatial relationships, plan workflows, and execute geospatial operations (Akinboyewa et al. 2025; Li, Ning, et al. 2025).

Although the users can benefit from LLMs to automate and get all their tasks way easier than the past, LLMs have significant drawbacks, including expensive computational resources (Lu et al. 2024). To address these issues in recent years SLMs were introduced. SLMs are emerging as efficient alternatives to large-scale models, particularly in scenarios with constrained computational resources (Lu et al. 2024). These models maintain competitive performance on narrow tasks while requiring fewer parameters, smaller memory footprints, and less training data. Their architecture is often optimized for inference on mobile devices or edge systems, making them highly relevant for decentralized applications (Javaheripi et al. 2023; Schick and Schütze 2020; P. Zhang et al. 2024). The emphasis is not merely on size reduction but also on task-specific adaptability through efficient fine-tuning techniques (Magister et al. 2022).

Several SLMs have gained popularity for their efficiency and performance on various NLP tasks. T5 (Text-To-Text Transfer Transformer), developed by Google, reframes all NLP tasks as text-to-text problems and is available in smaller versions like T5-small model for lightweight applications (Raffel et al. 2020). DistilBERT, a distilled version of BERT, retains 97% of BERT's language understanding capabilities while being 60% faster and significantly smaller (Sanh et al. 2020). ALBERT (A Lite BERT) reduces model size by sharing parameters across layers and employing factorized embeddings, making it more memory-efficient without sacrificing much performance (Lan et al. 2019). MiniLM is another compact model that achieves strong results on benchmark tasks using deep self-attention distillation techniques, offering a good trade-off between speed and accuracy (W. Wang et al. 2020b). These models are widely used in environments with limited computational resources. The integration of SLMs into GIS is rapidly gaining attention as a practical and efficient solution for enhancing spatial data understanding, automation, and user interaction. Unlike large-scale LLMs, SLMs offer computationally lightweight alternatives that can be fine-tuned for domain-specific geospatial tasks. For example, SpaBERT introduces geographic coordinate embeddings into BERT, significantly improving geo-entity typing and linking accuracy by encoding spatial relationships directly into the language model architecture (Li et al. 2022).

Having mentioned their capabilities, SLMs present an encouraging answer to the computational demands normally experienced in geospatial applications, especially where real-time edge processing is necessary. By placing SLMs directly on users' devices, such as within a web browser, developers can enable local interpretation of queries, geospatial tagging, and spatial pattern recognition without sending data to centralized cloud infrastructure. The distributed strategy improves user privacy, diminishes network dependency, and makes responsive functionality possible in bandwidth-limited or offline settings, e.g., in agriculture, transportation, or disaster response use cases. The primary challenge lies in the development of domain-specific datasets and fine-tuning strategies that map lightweight models to the spatial reasoning tasks needed for precise GIS outputs. This research investigates and compares various fine-tuning methods for SLMs and evaluates their usefulness in deploying autonomous GIS features on low-resource client-side environments. The main objective is to develop an efficient framework with the ability to execute geospatial tasks automatically in a distributed environment.

## 3. Materials and method

In this study, three autonomous GIS approaches were implemented and evaluated based on their effectiveness in natural language translation into geospatial function calls in order to automate Web GIS tasks. A single unified dataset was created, which was applicable to all three approaches. This guarantees a fair comparison among models. The data preparation process and modeling methodologies are explained in detail in the following sections.

### 3.1. Data preparation

Each human-generated text was labeled with the corresponding GIS function and its related parameters. In order to prepare the training dataset, first, natural language queries were collected for each target GIS function. ChatGPT-4 and Cohere each generated 100 examples per function, resulting in a total of 200 samples for every function. Subsequently, redundant or near-duplicate entries were removed across both sources. After cleaning and merging, the dataset contained approximately 2,000 unique samples.

All samples are assigned to their corresponding GIS function and required parameters. This ensures alignment with the input-output structure that text-to-text models expect. A random partitioning was done to split the dataset into subsets with an 80% and 20% proportion, making up the train and test datasets, respectively. Table 1 provides a list of all the GIS functions used in this research. To guide the generation process, the general prompt shown in Figure 1 is used.

Table 1. The distinct function call types/classes present in the data.

| Function | Description |
|---|---|
| AddMarker | Adds a marker (point of interest) to the map at specified coordinates, often with a label or name. |
| AddLayer | Adds a new map layer (such as a base map or thematic layer) to the map display. |
| AddVector | Loads and displays a vector data file (such as point, polyline, or polygon features) on the map. |
| AddWMS | Adds a map layer from a Web Map Service (WMS) URL, allowing integration of remote geospatial data. |
| Cartography | Changes cartographic properties of the map, such as background color, fill color, or stroke style. |
| Draw | Initiates a drawing action on the map, such as drawing a point, line, or polygon. |
| Move | Moves or pans the map view to center on specified coordinates. |
| MoveToExtent | Adjusts the map view to fit a specified bounding box or extent, defined by two sets of coordinates. |

| ZoomIn | Zooms the map in by a specified number of levels, making the view more detailed. |
| --- | --- |
| ZoomOut | Zooms the map out by a specified number of levels, making the view less detailed and showing a larger area. |

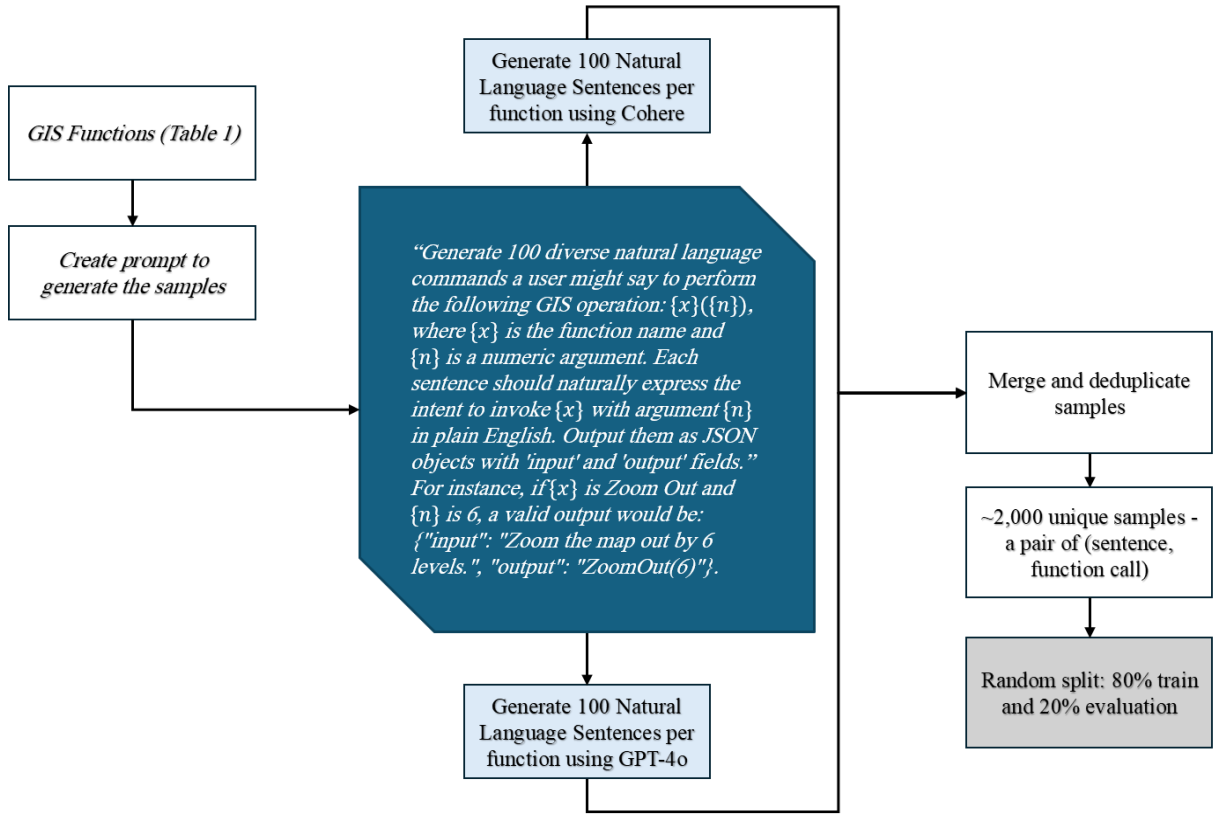

Figure 1. Overview of the dataset creation process for GIS function modeling

## 3.2. Modeling

This study contains approaches designed to automatically perform GIS operations inside the client-side browser environment (Figure 2). When a geospatial task is requested by the user, the corresponding function cell is identified by the application. Transforming inputs, which are in natural language, into relevant function calls executed on the client side is the main goal of all three approaches. The following sections describe each approach separately in detail.

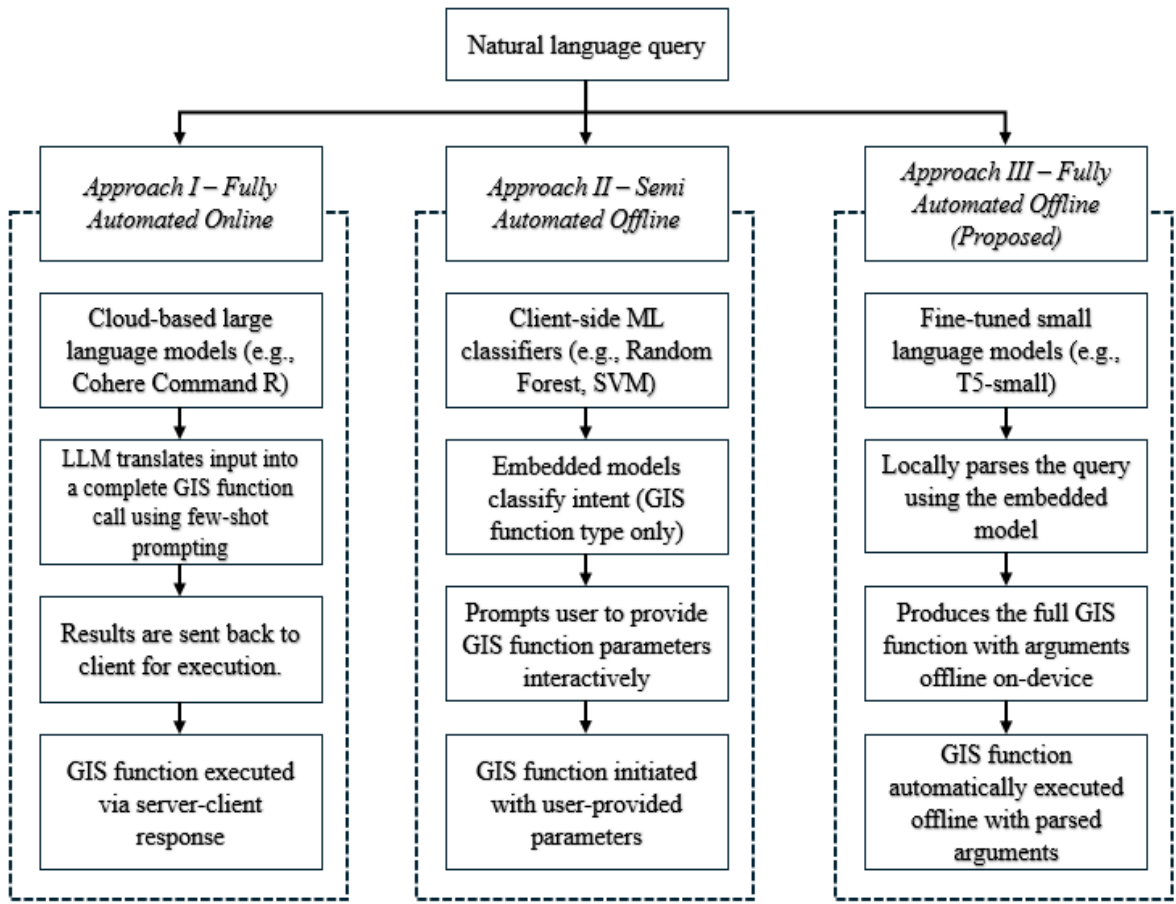

Figure 2. Conceptual workflow of the three autonomous web-based GIS (AWebGIS) approaches

### 3.2.1. Approach I: autonomous web-based GIS (AWebGIS) using LLMs

In this approach, the geospatial operations on the Web GIS application are conducted based on queries input by users in natural language. Using Hypertext Transfer Protocol (HTTP), these requests are transmitted to the server (Fielding and Reschke 2014), which has the responsibility of translating natural language into executable function calls using LLMs in order to understand the purpose of the user. To achieve this goal, few-shot learning is utilized to provide prompt examples for the LLM prior to actual deployment.

In this research, tailored prompts were constructed to instruct LLM in translating user queries into corresponding geospatial function calls. For the translation tasks, the Cohere Labs Command R 08-2024 language model (Cohere Labs 2024) was employed. This model is a 32-billion parameter generative LLM, optimized for reasoning, summarization, and function calling, and is well-suited for geospatial query translation (Cohere Labs 2024). Additionally, temperature was set to 0 to reduce random generation of outputs and the maximum number of tokens generated by model to 64 tokens for balancing the number of output tokens generated by the model. Eventually, the prompt shown in Figure 3 was used to guide the LLM (few-shot learning).

*"You are an expert system that translates user queries into geospatial function calls. Here are some examples:\nUser: I'd like to zoom out by 2 levels\nFunction Call: ZoomOut(2)\nUser: Show the seismic activity map from WMS URL https://example.activity/wms\nFunction Call: AddWMS('https://example.activity/wms')\nUser: Load the point vector using point_zones_NY_kpn.kml!\nFunction Call: AddVector('point', 'point_zones_NY_kpn.kml')\nUser: Add marker 'University' at location -73.1888, 122.889!\nFunction Call: AddMarker('University', [-73.1888, 122.889])\nUser: Set map bounds from 62.2585, -120.3652 to 63.8833, -3.3906.\nFunction Call: MoveToExtent(62.2585, -120.3652, 63.8833, -3.3906)\nUser: Switch to the OpenMallMap layer for retail therapy.\nFunction Call: AddLayer('OpenMallMap')\nUser: Can we go to 40.5267, -79.4892?\nFunction Call: Move(40.5267, -79.4892)\nUser: Draw a Line on the map!\nFunction Call: Draw('Line')\nUser: Set the background color to ivory.\nFunction Call: Cartography('background', 'ivory', null)\nUser: Zoom in by 7 levels to focus on the details.\nFunction Call: ZoomIn(7)\nUser: {User Query}\nFunction Call:"*

Figure 3. The prompt for few-shot learning of the Cohere model to predict function calls from user's queries

Additionally, three evaluation metrics were employed to assess the effectiveness of the model in this setup:

### (1) Exact match accuracy (EMA)

This evaluation metric measures the proportion of predictions that exactly match the ground truth output, as defined in Eq.1. The entire predicted output must be an exact match to the expected target, which makes this metric a strict one.

$$EMA = \frac{Number\ of\ exact\ matches}{Total\ number\ of\ samples} \tag{1}$$

*(2) Levenshtein similarity (LS)*

Levenshtein similarity measures how similar two strings are based on the minimum number of single-character edits (insertions, deletions, or substitutions) required to change one string into the other. The similarity score is computed as Eq.2 (Berger, Waterman, and Yu 2020; Po 2020; S. Zhang, Hu, and Bian 2017):

$$LS = 1 - \frac{D(s1,s2)}{max(|s1|,|s2|)} \tag{2}$$

where $D(s1,s2)$ is the Levenshtein distance between strings $s1$ and $s2$, and $|s|$ is the length of the string s.

*(3) Recall-oriented understudy for gisting evaluation (ROUGE) scores*

*ROUGE* is a set of metrics for evaluating automatic summarization. *ROUGE-1* considers unigram overlap, while *ROUGE-L* considers the longest common subsequence (Kotkar et al. 2024):

$$ROUG - 1 = \frac{Number\ of\ overlapping\ unigrams}{Total\ unigrams\ in\ reference} \tag{3}$$

$$ROUG - L = \frac{LCS\ (reference,\ candidate)}{length\ of\ reference} \tag{4}$$

In Eq.4 $LCS$ is the length of the longest common subsequence between the candidate and the reference summary.

### 3.2.2. Approach II: A WebGIS using traditional NLP models (semi-automated)

The second approach focuses specifically on identifying the type of function provided by the user, without performing parameter extraction. In this scenario, traditional ML models, which are designed for classifying input prompts into a list of possible function calls, are pre-trained and integrated directly into the user's browser.

Once the inputs are classified and the function call is identified (by the ML model), a pop-up input box is dynamically generated by the application and then presented to the user to ask for the necessary input parameters. random forest (RF) (Breiman 2001) and SVM (Cortes and Vapnik 1995; Hearst et al. 1998) are two of these classical ML models, which were trained on the dataset described earlier.

*(1) SVM*

SVM introduced by (Cortes and Vapnik 1995), are supervised learning models designed for classification tasks. The idea behind SVM is to find an optimal hyperplane that separates data points of different classes with the maximum margin. This is done by mapping the original input space into a high-dimensional feature space using a nonlinear transformation, where a linear decision boundary is constructed (Cortes and Vapnik 1995).

Mathematically, given a set of training examples $(x_i, y_i)$ with labels $y_i \in \{-1, +1\}$, the SVM aims to find a hyperplane expressed by Eq.5. w is the weight vector (normal to the separating hyperplane) that determines its orientation, and b is the bias term that shifts the hyperplane in space (Cortes and Vapnik 1995).

$$f(x) = sign(w.x + b) \tag{5}$$

Such that it maximally separates the two classes. The optimal hyperplane is the one that maximizes the margin $\frac{2}{\|w\|}$, and is found by solving a convex quadratic optimization problem of Eq.6 (Cortes and Vapnik 1995).

$$\min \frac{1}{2}\|w\|^2 \quad subject\ to \quad y_i(w.x_i + b) \geq 1 \tag{6}$$

For non-linearly separable data, SVM introduce slack variables and a regularization parameter $C$ to allow for some misclassification (soft-margin SVM), minimizing the Eq.7.

Also, slack variables $\xi_i \geq 0$ are introduced to measure how much each training point violates the margin (Cortes and Vapnik 1995).

$$\frac{1}{2}\|w\|^2 + C \sum \xi_i \quad subject\ to \quad y_i(w.x_i + b) \geq 1 - \xi_i , \xi_i \geq 0 \tag{7}$$

A key innovation of SVM is the use of kernel functions $K(x_i, x_j)$, which compute the dot product in the high-dimensional feature space without explicitly transforming the data. This allows SVM to efficiently learn nonlinear decision boundaries using functions like polynomial kernel and radial basis function (RBF). Only a subset of the training data points — called support vectors — influence the final model. These points lie closest to the decision boundary and define the margin (Cortes and Vapnik 1995).

Known for their exceptional generalization performance, SVM can especially deal with problems where features outnumber training samples. Their formulation provides a balance between precise control over the complexity of the model and error, which is critical when handling GIS applications due to their high dimensionality (Andris, Cowen, and Wittenbach 2013).

In our experiments, we utilized LinearSVC classifier from the scikit-learn library that was applied with its default hyperparameter settings. The LinearSVC classifier, which implements a linear SVM, uses an L2 regularization and a squared hinge loss function by default. It is optimized using the dual formulation (dual=True), which is efficient when the number of features exceeds the number of samples. The regularization parameter is set to $C$ =1.0, which controls the trade-off between achieving a low training error and a low testing error. The model includes an intercept term (fit_intercept=True), uses a one-vs-rest strategy (multi_class='ovr') for multi-class problems, and is limited to 1000 optimization iterations with a convergence tolerance of 1e-4.

*(2) RF*

An ensemble learning method, RF, which was first introduced by (Breiman 2001), generates several decision trees and integrates their outputs to carry out classification or regression tasks. Its fundamental principle is that a group of weak learners can form a robust predictive model when combined together. The ultimate output of classification problems is decided based on the majority vote from all trees within the forest.

Each individual tree in a RF is trained on a bootstrap sample —a randomly selected subset of the training data with replacement. Moreover, at each node, only a randomly selected subset of features is considered when determining the optimal split. This approach introduces diversity among trees, which reduces the risk of overfitting and improves generalization (Breiman 2001).

Formally, given a training dataset $D = \{(x_1, y_1), (x_2, y_2), \ldots, (x_n, y_n)\}$, the algorithm proceeds as follows:

1. Draw $B$ bootstrap samples from $D$.

2. For each sample, train a decision tree. At each node, select the best split from a random subset of $m$ features (where $m < M$, the total number of features).

3. Aggregate the predictions of all $B$ trees using majority voting (for classification) or averaging (for regression).

Key advantages of RF include its robustness to noise and overfitting, achieved through the averaging of predictions across multiple uncorrelated trees. Additionally, it offers built-in mechanisms for estimating feature importance by measuring each feature's contribution to impurity reduction throughout the forest. RF is also highly scalable, as the training of individual trees can be performed in parallel, making it efficient for large datasets (Breiman 2001). These characteristics make RF particularly effective in high-dimensional settings and with heterogeneous data types, which are common in GIS tasks (Georganos et al. 2021).

In our experiments, we utilized RandomForestClassifier that constructed 100 decision trees by default (n_estimators=100). Each tree was trained using the Gini impurity criterion to evaluate the quality of splits, with no limit on tree depth (max_depth=None), allowing trees to grow until all leaves are pure. The model considered the square root of the total number of features when choosing the best split at each node (max_features='sqrt') and used bootstrap sampling (bootstrap=True) to create diverse subsets of the data for training each tree. Minimum sample thresholds for splits and leaves were set to min_samples_split=2 and min_samples_leaf=1, respectively.

The following metrics were computed on the test set to assess performance (Powers 2020):

$$Precision = \frac{TP}{TP+FP} \tag{8}$$

$$Recall = \text{Sensitivity} = \frac{TP}{TP+FN} \tag{9}$$

$$Accuracy = \frac{TP+TN}{TP+TN+FP+FN} \tag{10}$$

$$F1\ Score = \frac{2(Precision)(Recall)}{Precision+Recall} \tag{11}$$

*Precision*, *recall*, *accuracy*, and *F1 Score* are derived from the confusion matrix, which includes four types of outcomes: true positives (*TP*), false positives (*FP*), false negatives (*FN*), and true negatives (*TN*). A *TP* occurs when the model correctly predicts a positive instance (e.g., detecting a relevant class when it is actually present). A *FP* is when the model incorrectly predicts a positive outcome for a case that is actually negative. Conversely, a *FN* means the model fails to detect a positive instance, labeling it as negative. A *TN* refers to correctly identifying a negative instance as negative. These four values are the basis of the evaluation metrics: *precision* reflects how many of the predicted positives were actually correct. *Recall*, or *sensitivity* indicates the model's ability to capture all actual positives. *Accuracy* considers both correct positives and negatives, reflects the overall correctness of the model. Lastly, the *F1 Score*, balances *precision* and *recall*, making it especially useful in situations with class imbalance where accuracy alone may be misleading (Powers 2020).

### 3.2.3. Approach III: Proposed method – AWebGIS using SLMs

The limitations of the first and second methods are tackled in this approach. In order to build an AWebGIS, the first method demanded a constant internet connection and was dependent on online services. Although able to operate offline, the second approach was only semi-automated. It was restricted mainly to classification tasks and was unable to summarize text and predict function calls alongside input variables. However, our proposed provides an effective solution to these problems.

In this approach, the T5-small model is utilized. In order to achieve real-time, client-side inference without requiring server communication, this model is entirely implemented within the browser environment. This design is very well-suited for interactive web-based applications due to its offline availability, low latency, and strengthened user privacy. Our method highlights an optimal trade-off between the compactness of the model and output quality, ensuring it remains practical for real-world applications where computational resources are limited.

T5-small model due to its architecture, deals with every natural language problem as a text-to-text one, whether the task involves translation, classification, answering question or summarization. This unified framework, together with its encoder-decoder design and multi-head self-attention mechanisms, provides the model with a strong ability to adapt and generalize effectively across a wide range of tasks (Raffel et al. 2020).

Our implementation employed a transfer learning approach based on the T5-small model (Raffel et al. 2020), integrated with a comprehensive data preparation pipeline. The test set was strategically split into validation (50%) and test (50%) subsets. Tokenization was performed using the T5-small tokenizer with a maximum sequence length of 64 tokens and the 'max_length' padding strategy, ensuring uniform input formatting across all samples (Hugging Face 2025a).

The training configuration included 20 epochs, a batch size of 8 for both training and evaluation phases, and the AdamW optimizer (Loshchilov and Hutter 2019) combined with the default T5 learning rate scheduler (Hugging Face 2025b). Model evaluation relied on two primary metrics: *EMA* for assessing perfect matches between predicted and reference outputs, and *LS* to evaluate normalized string similarity. The final model was exported in Open Neural Network Exchange (ONNX) format for future deployment within browser environments (ONNX Runtime developers 2021).

### 3.2.4. Web application development

As a core contribution of this study, we developed a demo web application to showcase three distinct approaches for translating natural language geospatial queries into actionable map operations within a browser-based GIS environment. The frontend is built with React and TypeScript, utilizing the OpenLayers library for interactive map rendering and spatial operations (The OpenLayers Dev Team 2025).

Based on Figure 4, for online inference, the application integrates the Cohere API to generate function calls from user queries in real time (Cohere 2024). The offline semi-automated mode instead uses ONNX models—including SVM and RF classifiers—executed entirely in the browser via the onnxruntime-web library (ONNX Runtime developers 2021). This dual-mode setup supports both cloud-based and fully client-side inference, enabling flexible deployment and improved user privacy.

The fully offline mode relies on the transformers.js library (Hugging Face 2024), which enables in-browser execution of transformer-based models. In this configuration, a T5-small model, fine-tuned for geospatial command translation, is loaded and executed locally. Tokenization and inference are performed entirely in the browser, with no communication with external servers.

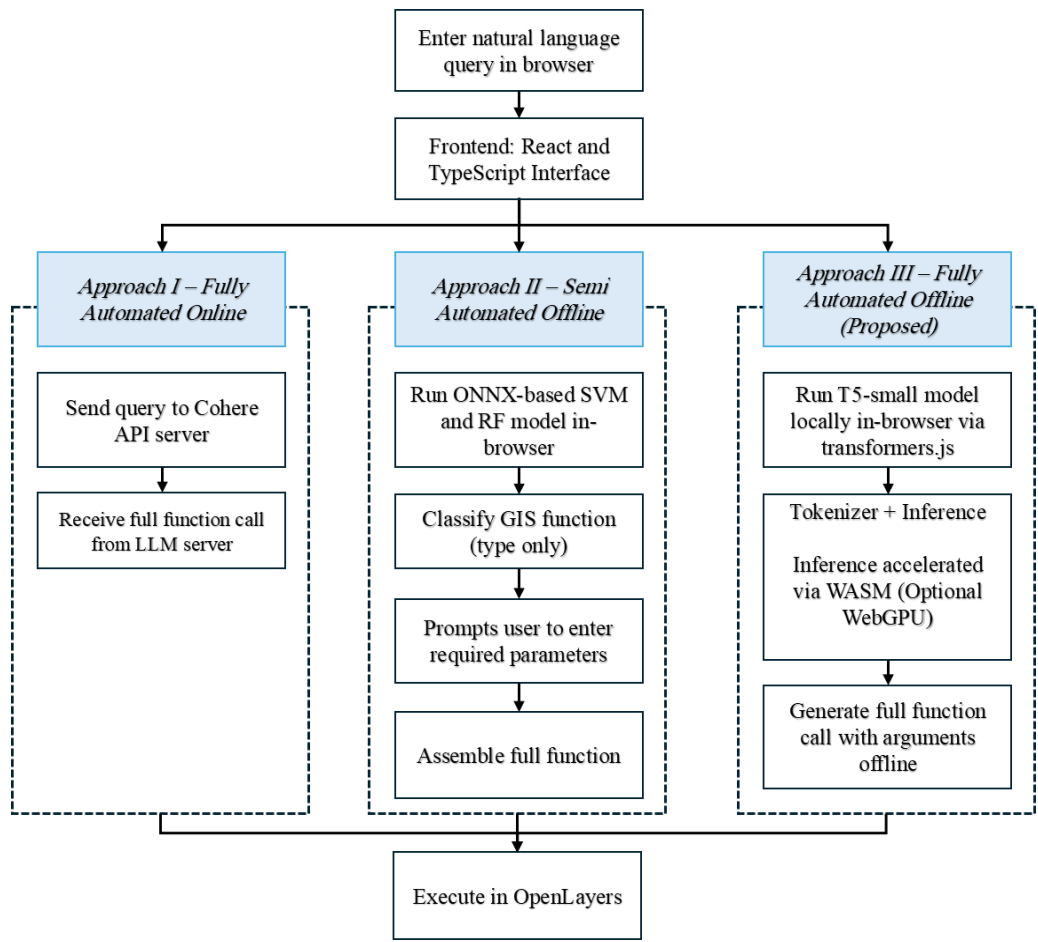

Figure 4. Workflow illustrating the process from prompt input to the execution of geospatial tasks across the three studied approaches

To support efficient execution of ML models in the browser, the application uses WebAssembly (WASM) as the default backend runtime. WebAssembly is a low-level, binary instruction format designed to run at near-native speed and provide a safe, portable, and efficient compilation target for modern web applications (WebAssembly Working Group and Rossberg 2025). WASM allows performance-critical tasks—such as neural network inference—to run within the browser environment with significantly better performance than JavaScript alone.

Furthermore, users can opt to switch to WebGPU mode via transformers.js, which leverages the browser's GPU to accelerate computations. This enables even faster inference by offloading matrix operations to the GPU, provided the user's browser and hardware support WebGPU (GPU for the Web Working Group et al. 2025).

Additionally, the transformers.js library includes an optimized decoder architecture that supports efficient sequence generation using a key-value caching mechanism. In standard transformer-based generation, the first decoding step is performed without any cache, while subsequent steps can leverage previously computed key and value tensors to avoid redundant computations. To streamline this process, transformers.js employs a merged decoder model that intelligently switches between the non-cached and cached decoding modes internally. This unified design also simplifies deployment and improves performance during inference, particularly in web environments where minimizing latency and memory usage is critical (Hugging Face 2024).

## 4. Experimental results

In this research, we employed three different methods to create an AWebGIS application and compared their efficiency based on certain evaluation metrics. Following the training of the models with few-shot learning (Approach I), supervised learning (Approach II), and fine-tuning (Approach III), we tested their performance on the same test set. Table 2 illustrates the accuracy for all the models employed in this research.

Based on Table 2, several models were assessed for AWebGIS development, each evaluated using task-appropriate performance metrics. In the online text-summarization category (Approach I), Cohere Labs Command R 08-2024 (a 32-billion parameter model) achieved an *EMA* score of 0.77 and an *LS* score of 0.93, along *ROUGE-1* and *ROUGE-L* scores of 0.91 for both of them. *Precision*, *recall*, and *F1 Score* were not reported for this model.

In the offline classification category (Approach II), the SVM model (~150 Kilo Bytes) did not report *EMA*, *LS*, or *ROUGE* scores, but achieved perfect values of 1.00 for *precision*, *recall*, and *F1 Score*. Similarly, the RF model (~4 Mega Bytes) also did not report *EMA*, *LS*, *BLEU*, or *ROUGE* metrics, but obtained 0.98 for each of *precision*, *recall*, and *F1 Score*.

In the offline text-summarization category (Approach III), the T5-small Float32 model (60 million parameters, ~295 Mega Bytes) recorded an *EMA* of 0.93 and an *LS* of 0.99, alongside *ROUGE-1* and *ROUGE-L* scores of 0.98 and 0.98 respectively. *Precision*, *recall*, and *F1 Score* were not reported, as these metrics were not applicable to the summarization task.

Table 3 represents significant differences between the three approaches in terms of autonomy level, accuracy, and user privacy. Approach I shows high autonomy and accuracy. However, it reveals low user privacy due to its reliance on online LLM queries. Approach II offers high user privacy since it runs offline. Nevertheless, low autonomy and accuracy (limited to basic classification tasks) are the drawbacks of this approach. Approach III revealed more balanced results. It demonstrates high user privacy and accuracy through offline summarization techniques, while providing a high level of autonomy. This comparison shows that Approaches II and III may be better options for privacy-conscious environments, despite the better overall performance exhibited by Approach I and III. Therefore, Approach III offers a balance between performance and privacy.

Table 2. Accuracy of models used for AWebGIS development

| Approach | Type | Model | Size | *EMA* | *LS* | *ROUGE-1* | *ROUGE-L* | *Recall* | *Precision* | *F1 Score* |
|---|---|---|---|---|---|---|---|---|---|---|
| **I - Online** | Text-Summarization (Fully-automated) | Cohere Labs Command R 08-2024 | 32-billion parameters | 0.77 | 0.93 | 0.91 | 0.91 | - | - | - |
| **II - Offline** | Classification (Semi-automated) | SVM | ~150 KB | - | - | - | - | 1 | 1 | 1 |
| | | RF | ~ 4 MB | - | - | - | - | 0.98 | 0.98 | 0.98 |
| **III - Offline** | Text-Summarization (Fully-automated) | T5-small Float32 | 60 million parameters ~295 MB | **0.93** | **0.99** | **0.98** | **0.98** | - | - | - |

Table 3. Comparison of the three approaches in terms of user privacy, autonomy level, and accuracy for AWebGIS development

| Approach | User Privacy | Autonomy Level | Accuracy | Model's Size |
|---|---|---|---|---|
| **I** | Low | **High** | **High** | Large |
| **II** | **High** | Low | Low (limited to basic classification tasks) | **Small** |
| **III (Propose Approach)** | **High** | **High** | **High** | **Small** |

Figure 5, visually compares three approaches (I, II, and III) across four key attributes: user privacy, autonomy level, accuracy, and model size. Each attribute is assessed qualitatively on a relative scale (e.g., low, moderate, high) based on performance characteristics observed in our experiments and system analysis. Approach I shows strong performance in autonomy and accuracy but scores poorly on user privacy and model size due to its large, online-dependent architecture. Approach II performs best in user privacy and model size, as it is lightweight and fully offline, but its autonomy and accuracy are limited to basic classification tasks, making it less suitable for complex scenarios. Approach III demonstrates the most balanced performance across all four attributes. It achieves high scores in autonomy level, accuracy, and user privacy, while also maintaining a reasonably small model size.

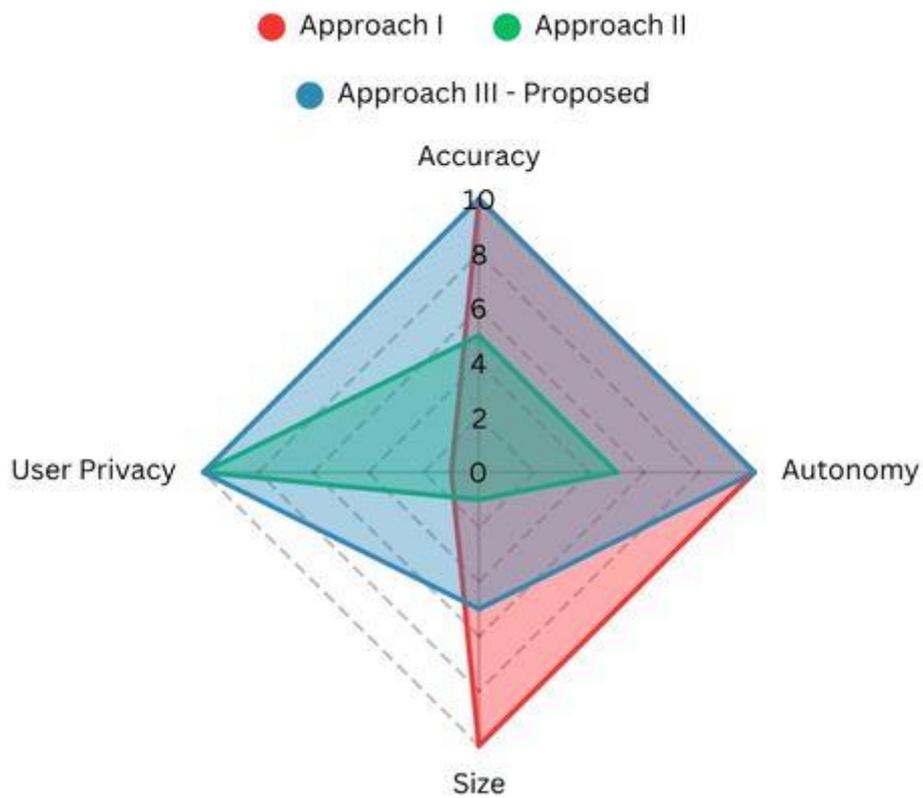

Figure 5. Comparison of three approaches based on four key attributes: user privacy, autonomy level, accuracy, and model size

Based on Figure 6, in Approach I, we employed a Cohere API as well as few-shot learning. The sentence "Show a marker at -4.5, 40 with Madrid as label" sent to the Cohere API then it was translated to a function call to show a new marker in Madrid.

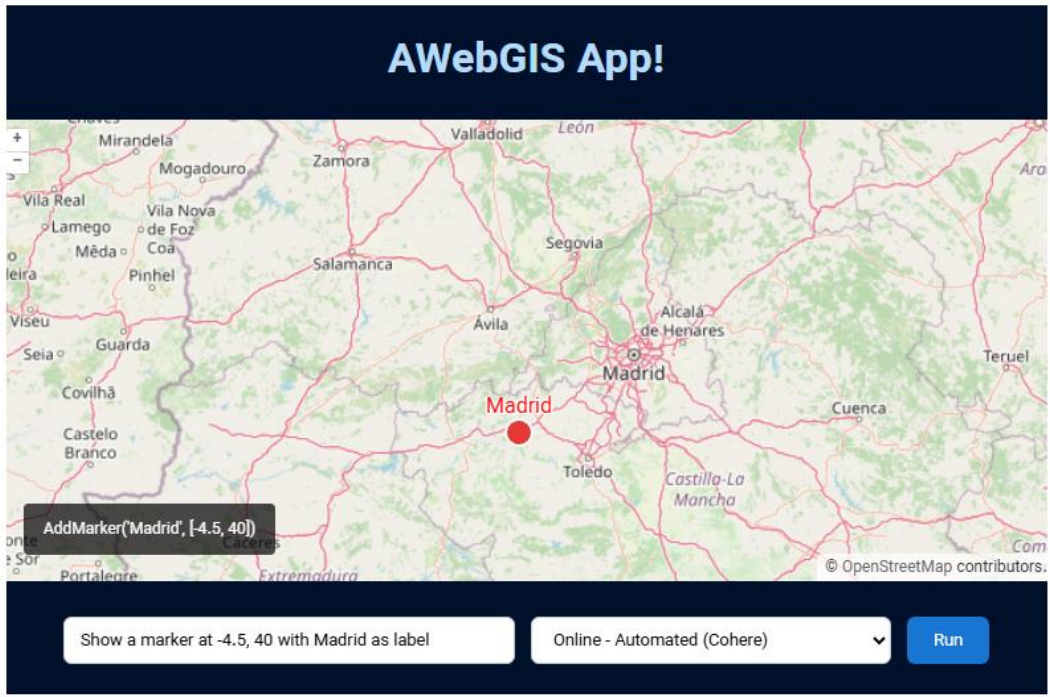

Figure 6. Using Cohere API to translate user's sentence to function calls

Based on Figure 7, in Approach II, we employed a semi-supervised learning strategy. Since this approach treats the problem as a classification task and does not use ML models to predict function parameters directly, the application retrieves these parameters from the user after the classification step. This design maintains user involvement while simplifying the model's predictive responsibilities.

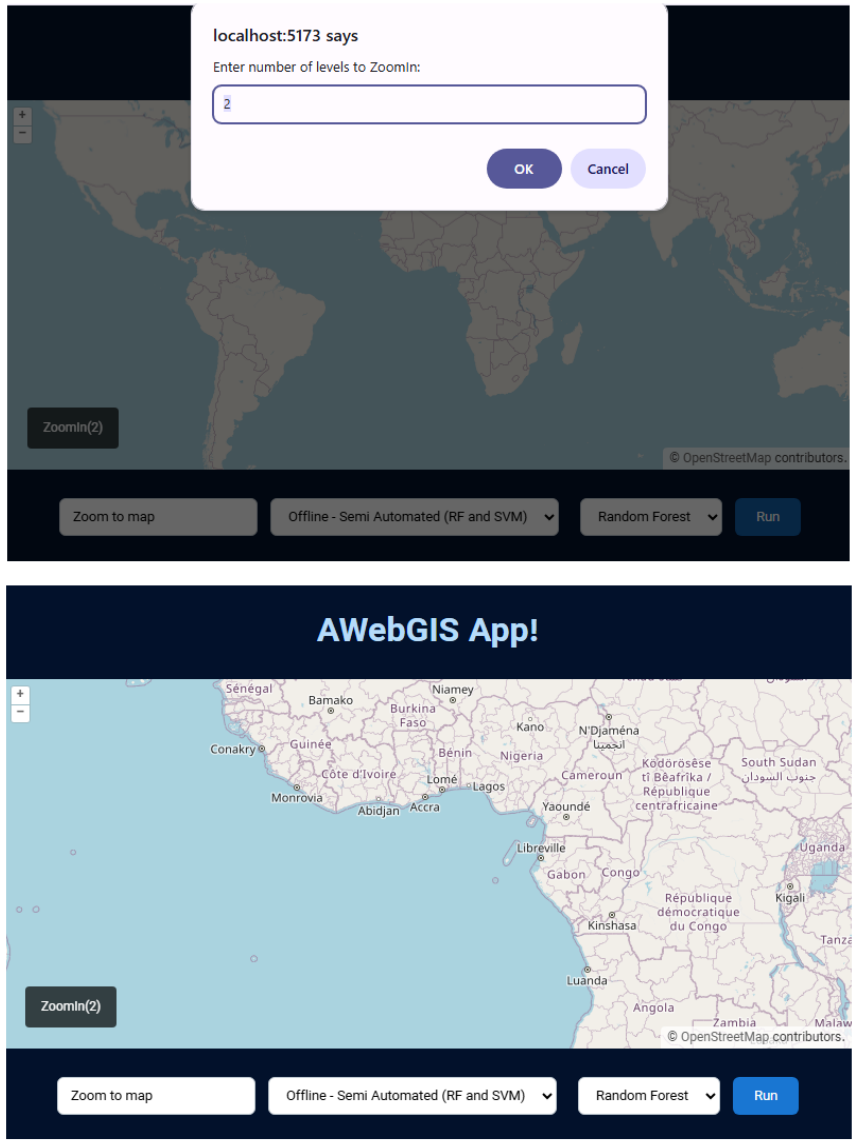

Figure 7. Classify user's prompt using RF model on the web browser

Based on Figure 8, in Approach III, we predicted function calls using the fine-tuned model based on T5-small model. The sentence "Show marker at -9.5, 39 'Portugal' is label" sent to the model and after prediction on the browser using WASM, the command was translated to a function call to show a new marker in Portugal.

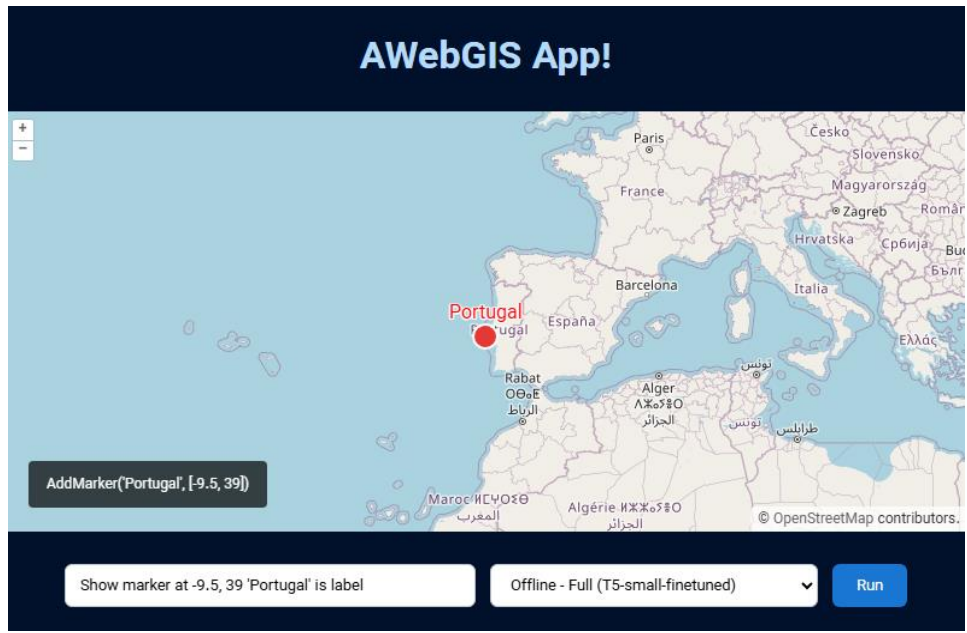

Figure 8. Using fine-tuned T5-small model to translate user's command into function calls

## 5. Discussion

### 5.1. Comparing approaches

The evaluation of three distinct architectural approaches for Autonomous Web-based Geographical Information Systems (AWebGIS) provides crucial insights into the trade-offs between performance, autonomy, and user privacy. Our findings demonstrate that each method—the online LLM, the offline classical ML, and the offline fine-tuned SLM—occupies a unique position in the AWebGIS design space.

The fully automated online approach, leveraging cloud-based LLMs like Cohere through few-shot learning, offered significant flexibility and high performance for a wide array of user instructions. This aligns with recent advancements in the field, where LLMs are increasingly used for complex geospatial tasks. For instance, studies by (Akinboyewa et al. 2025) and (Li and Ning 2023) have focused on benchmarking the capabilities of large, cloud-based models for multi-step geospatial operations. These works highlight the impressive potential of LLMs to interpret natural language, plan workflows, and execute complex GIS functions. However, as our study's results for this approach confirm, these solutions are inherently tied to continuous internet connectivity and raise significant concerns regarding user privacy due to the transmission of sensitive data to centralized servers.

In contrast, our semi-automated offline method, utilizing classical classifiers such as SVM and RF, effectively mitigates these privacy and connectivity issues. This approach classifies user intent and identifies the appropriate GIS function on the client side, but its limited automation and inability to handle complex, multi-argument queries restrict its utility. It represents a functional, but less flexible, solution for simple tasks in disconnected environments.

Our proposed third approach—a fully autonomous, client-side model using a fine-tuned T5-small language model—emerges as a compelling and novel alternative that addresses the core limitations of the other methods. By operating entirely within the user's browser, it eliminates the need for server-based inference, thereby guaranteeing user privacy and enabling a true hands-free, offline experience. The high accuracy of this approach (EMA of 0.93, Levenshtein similarity of 0.99, and ROUGE-L of 0.98) is a key finding, demonstrating that a lightweight, fine-tuned SLM can achieve performance on par with or even surpass its online LLM counterparts for the specific task of converting natural language queries into function-argument pairs. This finding is particularly significant when compared to studies that focus solely on the performance of large models, such as those (Wei et al. 2025) or (Akinboyewa et al. 2025), as it proves that a "small but mighty" approach is a viable path forward for domain-specific applications. The success of this method supports the growing trend of using SLMs for resource-constrained environments, as discussed by (Richard and Villanueva 2025), and aligns with research on the efficiency and on-device capabilities of models like T5-small model and other distilled models such as DistilBERT (Richard and Villanueva 2025). Our work provides a direct empirical comparison to high-resource online solutions by presenting a viable, high-accuracy offline alternative—a less explored but critical direction for real-world GIS deployments, especially in edge environments.

## 5.2. Limitations and future research

While our study highlights the potential of client-side SLMs for AWebGIS, it also reveals several limitations that warrant future research. First, the current T5-small model was fine-tuned on a specific dataset of 2,000 queries, limiting its ability to recognize a broad range of GIS functions and parameters. The operations primarily focus on a narrow set of tasks, leaving out crucial areas such as data management, advanced visualization, and complex spatio-temporal analysis. Future work could address this by expanding the training dataset to cover a wider spectrum of GIS functions and by exploring other powerful SLMs, such as Qwen2 (Q. Team 2024), Llama 3.1 (Touvron et al. 2024), Mistral Small (AI 2023), Phi-3 (M. Research 2024), SmolLM2 (Q. A. Research 2024), and MiniLM (W. Wang et al. 2020a), which have shown strong performance on various text-to-text tasks.

Second, the study employed a single fine-tuning strategy without exploring alternative knowledge distillation techniques (Gou et al. 2021). Methods such as Low-Rank Adaptation (LoRA) or QLoRA (Storz 2025) could be investigated to further enhance the model's efficiency and accuracy while reducing memory footprint.

Third, the current system lacks a memory or state-management mechanism, meaning it cannot handle follow-up questions or conversational context. This is a significant limitation for creating a truly intuitive user experience. Future research should explore the integration of client-side Retrieval-Augmented Generation (RAG) capabilities and memory caching (Z. J. Wang and Chau 2024; Fan et al. 2025). By implementing a vector database in the browser, the system could retrieve relevant information from a predefined knowledge base of GIS documentation to augment the model's understanding and provide more accurate and context-aware responses without sending user data to the cloud. This would combine the benefits of an SLM with the power of an external knowledge source, a technique gaining traction in various on-device applications.

Despite these limitations, this study successfully demonstrates the feasibility and high potential of leveraging fine-tuned, client-side SLMs for AWebGIS development. It lays a solid foundation for future work on distributed, privacy-preserving, and highly efficient geospatial systems. The approach presented here is a critical step towards democratizing access to powerful GIS tools by making them available to users with limited connectivity and hardware, which is a major need for various use cases, including disaster response, field work, and off-grid scenarios.

## 6. Conclusion

This study focuses on comparing three distinct language-driven Web GIS automation methods and evaluating them based on their autonomy, accuracy, computational performance, and user privacy. High accuracy and strong flexibility are provided by the fully automated online approach. However, it is extremely reliant on cloud infrastructure and compromises user privacy. The semi-automated offline approach offers lightness and preservation of privacy, while its automation is limited, and its dependence on classification techniques lowers its accuracy.

Our proposed fully automated offline approach, leveraging a fine-tuned T5-small model, emerged as a balanced solution—achieving high accuracy, maintaining user privacy, and operating efficiently without internet connectivity. Although it currently supports a limited set of GIS operations and uses only one model tuning strategy, it lays the groundwork for more advanced, edge-compatible GIS systems. Future work will expand the functional scope, explore other SLMs, and incorporate techniques like knowledge distillation, memory caching, and RAG to enhance autonomy and responsiveness.

*CRediT authorship contribution statement*

**Mahdi Nazari Ashani**: Conceptualization, Methodology, Validation, Data Curation, Writing – Original Draft, Writing – Review & Editing, Visualization, Supervision. **Ali Asghar Alesheikh:** Validation, Writing – Review & Editing, Supervision. **Saba Kazemi**: Methodology, Data Curation, Writing – Review & Editing, Visualization. **Kimya Kheirkhah**: Writing – Review & Editing, Visualization. **Yasin Mohammadi**: Methodology. **Fatemeh Rezaie**: Writing – Review & Editing, **Amir Mahdi Manafi**: Methodology. **Hedieh Zarkesh**: Methodology.

*Data availability*

The datasets used and/or analyzed in this study are publicly available. The code and related materials are accessible at **https://github.com/mahdin75/awebgis**, and the fine-tuned T5-small model can be found at **https://huggingface.co/mahdin75/awebgis**.